# A robust low data solution: dimension prediction of semiconductor nanorods

Xiaoli Liu [1†], Yang Xu [2†], Jiali Li [1], Xuanwei Ong [3], Salwa Ali Ibrahim [3], Tonio Buonassisi [4] and Xiaonan Wang [1]*

[1] Department of Chemical and Biomolecular Engineering, Faculty of Engineering, National University of Singapore, Block E5, Engineering Drive 4, 117585 Singapore

[2] Institute of Materials Research and Engineering, A*STAR 2 Fusionopolis Way, Innovis, #08-03 138634 Singapore

[3] Department of Chemistry, National University of Singapore Block S8 Level 3, 3 Science Drive 3 Singapore 117543

[4] Massachusetts Institute of Technology, 77 Massachusetts Avenue, Cambridge, MA 02139, USA

† These authors contribute equally.

* Corresponding author:

Tel: +65 6601 6221

Email: chewxia@nus.edu.sg

**Abstract**

Precise control over dimension of nanocrystals is critical to tune the properties for various applications. However, the traditional control through experimental optimization is slow, tedious and time consuming. Herein a robust deep neural network-based regression algorithm has been developed for precise prediction of length, width, and aspect ratios of semiconductor nanorods (NRs). Given there is limited experimental data available (28 samples), a Synthetic Minority Oversampling Technique for regression (SMOTE-REG) has been employed for the first time for data generation. Deep neural network is further applied to develop regression model which demonstrated the well performed prediction on both the original and generated data with a similar distribution. The prediction model is further validated with additional experimental data, showing accurate prediction results. Additionally, Local

Interpretable Model-Agnostic Explanations (LIME) is used to interpret the weight for each variable, which corresponds to its importance towards the target dimension, which is approved to be well correlated well with experimental observations.

**Introduction**

Upon decreasing the size of bulk materials to the nanometer regime, these nanomaterials begin to exhibit novel chemical, optical, electronic or magnetic characteristics that significantly differ from their bulk counterparts, resulting in applications ranging from optoelectronics and magnetic devices to catalysis and biomedicine.[1–4] In recent years, research focus on nanomaterials has been shifting from spherical nanomaterials to those with higher aspect ratios such as nanowires, nanorods, and nanotubes. Due to their elongated geometry, those nanomaterials usually possess larger surface area and anisotropic properties which offer new possibilities for various applications. For example, anisotropically shaped gold nanoparticles possess an additional tunable plasmon band towards the near-infrared (NIR) region compared to traditional spherical gold nanoparticles and enhanced surface enhanced Raman scattering (SERS) effect.[5] Anisotropic magnetic nanomaterials exhibited higher magnetic moments and aspect ratio dependent cellular uptake behavior which did not exist in the spherical magnetic nanoparticles.[6] Colloidal semiconductor nanocrystals, also known as quantum dots, are a well-studied class of nanomaterials with unique optoelectronic properties, resulting in applications in various fields such as photovoltaics, LED displays and biomedicines.[7–9] In recent years, the development of these colloidal semiconductor nanocrystals has also shifted towards anisotropic shaped nanocrystals.[10–14]

As one of the first few developed anisotropic semiconductor nanocrystals, CdSe-seeded CdS nanorods (NRs) have attracted tremendous interest for the past decade. Due to its anisotropic shape, such NRs exhibit unique properties compared to the typical spherical core-shell semiconductor nanocrystals (NCs).[15–20] Fluorescence from these NRs originates from the radiative recombination of excitons in the CdSe core while photon absorption at shorter wavelengths is dominated by the rod-like CdS shell.[15] Consequently, the absorption cross-section of NRs can be altered by changing the length of the NR while its emission wavelength exhibits little to no change.[21] This allows the NRs to have a much larger action cross section in principle when compared to spherical core-shell NCs, which corresponds to a much brighter particle with similar quantum yield. The aspect ratio of these NRs has also been shown to significantly affect their optical and electronic properties. Tunable degree of optical anisotropy is observed by tuning the diameter and length of the NRs, leading to a higher degree of linear polarization of the electronic transitions upon an increase in the shell thickness.[22] Additionally, different aspect ratios lead to different photoluminescence quantum yields and decay lifetimes, which are key parameters for various applications such as microscopy and sensing.[23] Aspect ratio has also been demonstrated to play an important role in photocatalytic applications due to changes in the dynamics of charge transfer processes.[24] Lastly, aspect ratio is also a key parameter when considering the self-assembly of these nanorods into mesostructures, affecting their usage in optoelectronic device applications.[25,26] Despite the significant improvement in synthesis and application of CdSe/CdS NRs, one major obstacle hindering its development is the laborious and time consuming experimental trial and error optimizations involved due to the complex behavior in the experimental parameter space. During the rod-like shell growth, the core size/amount, precursors amount, ligand type/ratio can impact the dimension of the rod and, in turn, optical properties of NRs. Thus, precise prediction of the dimension from specific conditions would accelerate material optimization process and help to control the resulting property in an application-oriented manner.

In recent years, machine learning (ML) methods have been widely applied in the fields of materials, chemical and biochemical engineering to solve various problems, such as material synthesis, drug generation, and genome analysis.[27–29] Recent advances in ML techniques, such as neural networks (NNs), and Bayesian optimization provide an unique opportunity for reshaping the optimization of nanocrystals, including perovskite[30,31], gold nanocluster[28] and PbS quantum dots.[32] Especially for material systems with limited understanding in synthetic processes, machine learning methods have brought efficient recipe prediction tools. However, in the field of materials science, the scarcity of material synthesis data and the high expense in acquiring them have hindered the application of ML in materials' synthesis understanding.[33,34] As shown by computer science communities, the key to successfully building an ML model is to learn rules and patterns from enough samples to effectively predict and explore unknown areas.[35,36] The literature on inorganic materials has demonstrated a strong correlation between predictive performance and sample size. For example, Faber et al.[37] developed a machine learning model that can systematically improve prediction accuracy as the size of the training set increases. Schmidt et al.[38] tested the machine learning benchmark method to predict the thermodynamic stability of solids. Through the construction of large-scale data training, it is found that the prediction error of most machine learning benchmark methods exhibits a monotonic power exponential decrement as the dataset increases. When the number of samples in the training set is doubled, the prediction error is reduced by about 20%. Ren et.al and Li et.al have tried to augment data by searching from literature sources or rely on simulated data. However, these methods either need simulated methods of the synthesis task which are not available for limited understood synthesis tasks or highly depend on the available public resources that can be scarce for new materials. It can be seen that the key to solving the problem is to reasonably generate samples for a small dataset based on statistical approaches to form an augmented one with sufficient samples that conforms to the distribution and fits the training model.

Herein, we have developed a deep neural network based regression algorithm for precise prediction of length, width, and aspect ratios of NRs. We employed the Synthetic Minority Oversampling Technique for regression (SMOTE-REG) for data generation up to 9,100 samples from 28 sample data obtained from bath synthesis. With the model developed, both the original dataset and the generated dataset were well predicted with a similar distribution, suggesting a well performed regression model with effective data augmentation. With the enriched training dataset, we further validated the model with additional experimental data, showing accurate prediction results. The weight for each experimental variable, which corresponds to its importance towards the target result, was further interpreted with the Local Interpretable Model-Agnostic Explanations (LIME). We found that the weights suggested by the model correlated well with experimental observations, which would serve as a guide for further experimental optimization.

**Results and Discussion**

**Data Collection**

CdSe/CdS nanorods were synthesised via hot injection seeded growth method.[10] We extracted the synthesis conditions of 28 examples, which produced monodisperse NRs with different widths, lengths, and aspect ratios. Each example contains three experimental variables, including (1) CdSe core absorption edge in wavelength (nm), which is representative of the core size; (2) CdSe core amount used for each reaction in nmol; (3) S precursor amount added during the CdS shell growth in mg. The average lengths and widths for each sample were calculated from 100 individual NR length and width measured from corresponding transmission electron microscopy (TEM) images using ImageJ software. The aspect ratios were independently calculated based on the average lengths and widths obtained for each nanorod sample. As illustrated by the random distributed points in Figure S1, the correlation of these three variables with average length, width, and aspect ratio is difficult to be explained by simple regression fitting.

**Data generation**

In order to increase the size and robustness of the original small data as training dataset, we first generated samples from the original 28 samples through the Synthetic Minority Oversampling Technique for regression. Generating new samples that match the original data distribution should significantly improve the predictive performance of the model trained. SMOTE (Synthetic Minority Oversampling Technique)[39,40] is a synthetic minority class oversampling method, which is widely used in unbalanced data classification problems. It can simulate data well and generate distribution-dependent neighbor samples for the minority class. SMOTE can flexibly generate the required percentage of samples for minority class, compared to the majority. However, SMOTE cannot be applied to regression data because regression problems predict continuous values rather than class labels. One key contribution of this work is to propose a strategy of SMOTE for addressing regression tasks with small samples. The key goals are: (1) accurately prediction of rare extreme values; and (2) discovery of new samples that match the original data distribution, which can be used to generate new ground truth data in experiments. For unbalanced classification, the original algorithm is based on information provided by the user about which class value is the minority class with fewer samples. In our problem, the target variable of continuous values needs to be processed. How to determine the target variable of the generated observations is the key issue we need to solve. In the original algorithm, this problem is easy to solve, because all samples of minority class have the same targets, and all the new samples generated on them belong to the same class. While for our regression task, the results are no longer straightforward. The scores of the over-sampled instances are highly correlated, but their target variable values should not be the same. This means that a new data point synthesized using a pair of samples may not have the same target variable value as them.

For the sake of overcoming the above limitations, a variant of SMOTE approach (SMOTE-REG) was proposed for regression task in this paper. We illustrated the strategy as follows, also shown in Figure 1, with detailed SMOTE-REG algorithm for dataset generation discussed in the section of Methods. As the size of our dataset was too small, we set the number of minority class samples to two, which meant that two samples were used as minority class, and the remaining samples were the other class. With any two samples from the original dataset as minority class, new data points were generated between the two, which were possible experimental conditions. In fact, for SMOTE-REG, the number of samples for minority class can be any value less than the total number of samples. When the number of samples of a minority class is greater than two, the generated samples are distributed among other samples in the same category neighborhood. Through this process, we obtained a new synthetic dataset with 9,100 samples. The new dataset synthesized is shown as blue dots in Figure 2, while the red dots representing original data. As demonstrated by the wider distribution of the generated data compared to the original one, the proposed method increases the diversity of generated samples covering a broad range of parameter space which would be difficult to obtain by experiment.

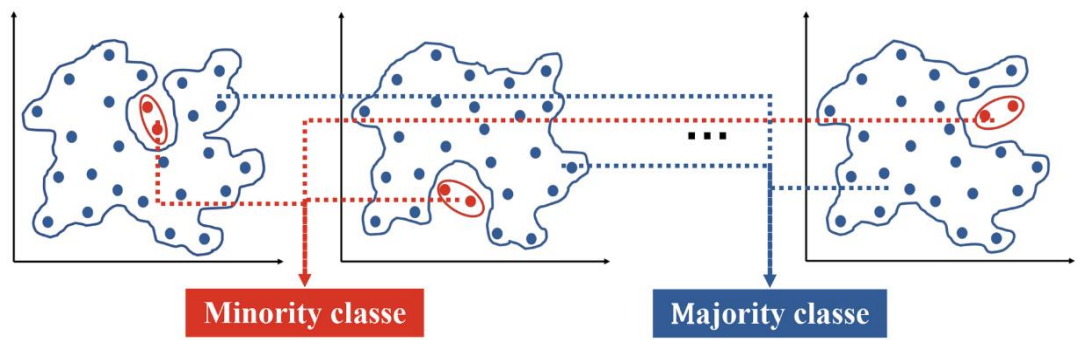

Figure 1 Regression data transformation.

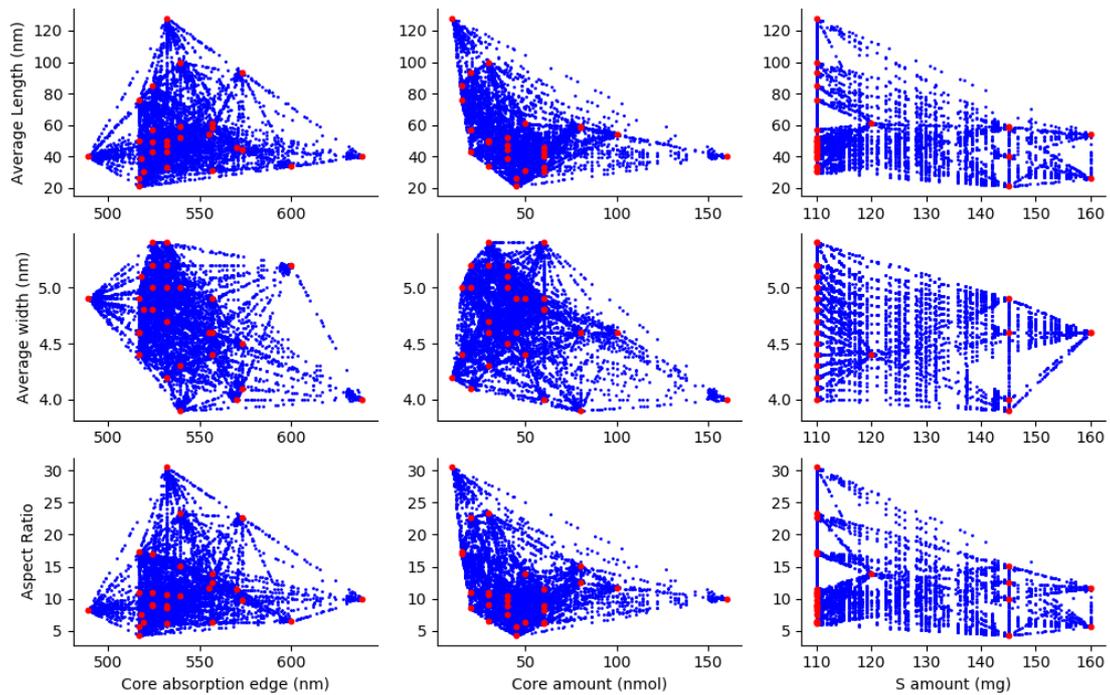

Figure 2 Data generated by SMOTE-REG.

## Regression Prediction

Based on synthetic samples, we proposed to use deep neural networks for regression prediction. A four-layer deep neural network was adopted that included an input layer, two hidden layers, and an output layer. The number of input layer nodes was three, corresponding to three experimental conditions. The number of two hidden layer nodes was set to 64 and the output layer was a single value. Three models were trained for predicting NR length, width, and aspect ratio, respectively. In order to adapt to the learning ability of different tasks, the learning rate was set to be 0.001, thus ensuring that the algorithm can converge stably. The data was z-scored before applying regression methods. For the quantitative performance evaluation, we employed the metrics of Mean Absolute Error (MAE), Mean Squared Error (MSE), and Coefficient of Determination ($R^2$) between the predicted scores and the target clinical scores for each regression task. Smaller values of MAE/MSE and larger values of $R^2$ indicate better regression performance. We evaluated the model by 10-fold cross-validation for the aggregated data, including both generated data and original data. The average and standard deviation of the performance measures were calculated by 10-fold cross-validation on different splits of data and summarized in Table 1. The evaluations of the original data (28 samples) are presented separately in the table. Figure 3(a), (c), and (e) show the scatter plots of actual values versus predicted values for the original raw data and the generated data, represented by red and blue dots, respectively. The results show that excellent prediction performance is achieved with deep neural networks in all tasks for the original dataset. This indicates that the data generated by our proposed SMOTE-REG method has a reasonably similar distribution to the original dataset, thus improving the training efficiency of the prediction models from the original small dataset. For the whole datasets, the prediction accuracy is not as good as the original dataset due to the wider distribution of generated data. The green and orange dots in Figure 3(b), (d) and (f) show the scatter plot of true values of the original sample and their predicted values, respectively, indicating that the predicted values are very close to the actual values for all three tasks across the entire samples.

To further validate the generated samples and accuracy of the prediction model. We used the generated data and original data as training set, predicting the length, width, and aspect ratio of additional three experimental samples given the three tasks mentioned previously. The validation samples were randomly selected with experimental conditions similar as the original data (see supporting Figure S3). As illustrated with the green and orange triangles in Figure 3(b), (d) and (f), all the three tasks were predicted with reasonable errors though the validation data were slightly out of the distribution of the training dataset. The last sample which fell closer in the distribution of the training dataset showed higher accuracy than the other two in all three tasks, further confirming the promising performance of the model developed in predicting NR dimensions in the vast parameter space covered by the enriched training dataset. As highlighted in the introduction, the properties of the NRs largely depend on their dimensions. With this prediction model developed, it is possible to potentially eliminate the time consuming experimental "try and error" optimization process when a desired dimension is required for a specific application, which will significant accelerate the process of unveiling the potential of NRs in virous areas.

Table 1 Results of DL on different scores by 10-fold cross validation.

|  |  | MAE | MSE | $R^2$ |
|---|---|---|---|---|
| Original data | Average Length (nm) | 2.7948 | 13.280 | 0.9761 |
|  | Average width (nm) | 0.0467 | 0.0038 | 0.9784 |
|  | Aspect Ratio | 0.6385 | 0.9289 | 0.9730 |
| Whole data including the generated and original (10-fold cross validation) | Average Length (nm) | 5.9316±0.2716 | 83.120±7.7793 | 0.7486±0.0325 |
|  | Average width (nm) | 0.1309±0.0049 | 0.0361±0.0028 | 0.5802±0.0376 |
|  | Aspect Ratio | 1.5703±0.0930 | 5.7627±0.6599 | 0.7055±0.0464 |

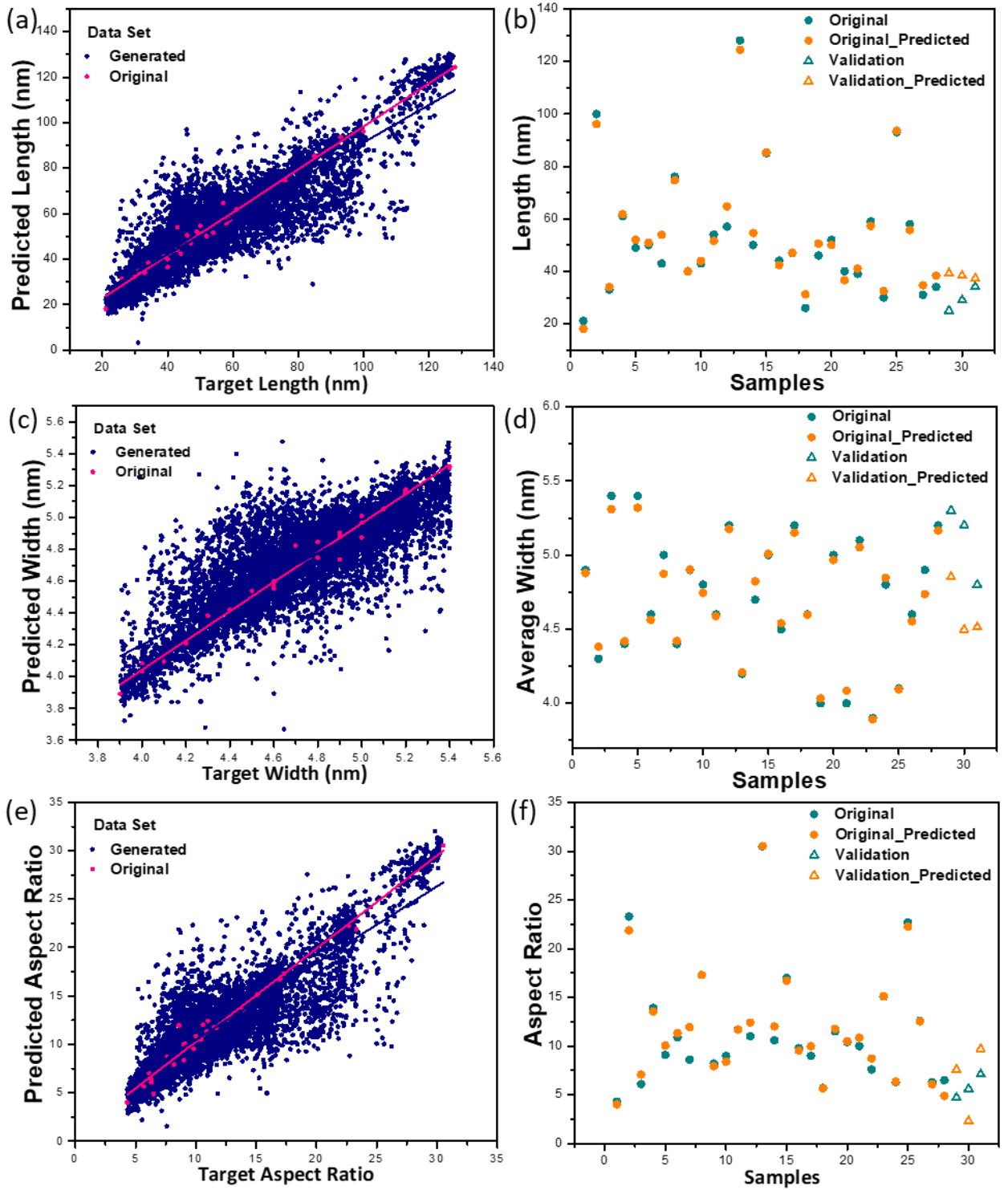

Figure 3 The correlation of experimental and predicted values by deep neural networks for length ((a) and (b)), width ((c) and (d)), and aspect ratio ((e) and (f)). The original data (red dots), generated data (blue dots) and their corresponding linear regression is shown at left column. The green and yellow dots (original) and triangles (validation) in right column represent the values from the experimental data and its corresponding predicted data

**Interpretation by Local Interpretable Model-Agnostic Explanations (LIME)**

To further illustrate the applicability of the model, we used LIME to interpret the results. LIME created a linear interpretable model within the neighborhood of each data point to be interpreted. The absolute value of the linear model weight can be used to indicate the importance of the feature. In our work, we used LIME to create a linear interpretation model for each sample. The interpretation results were expressed in the form of weighted features, which form a linear model to approximate the behavior of deep neural network model in the vicinity of the test example.

As shown in the weights heatmap for the three tasks in Figure 4, Core amount was generally more important than the other two features in determining length and aspect ratio. In the presence of the CdSe core seeds, CdS grew preferentially on it as there is lower activation energy for heterogeneous nucleation.[41] Thus, core amount in the reaction mixture determined the quantity of active surface site available for CdS nucleation and growth, affecting significantly the reaction kinetics, thermodynamics and final NR dimensions. For instance, comparing feature weights from sample 03, 11, 16, and 28, where the reactions started with the same core size and S amount while the core amount added varies, the importance of each reaction variable could be explained and matched with the modeling values. For sample 28 with the lowest amount of core added, the nucleation site was limited and a larger amount of precursor is available, resulting in fast reaction kinetics for anisotropic growth thus longer NRs obtained. Consistent with the feature weights, all the three features of sample 28 exhibited stronger correlation with the core amount then the rest 3 samples. When an increased core amount added in the samples from 28, 16, 11, to 03, slower growth kinetics was expected as more S consumed for nucleation, leading to weaker dependence of S amount in determination of NR length. In the case of sample 03 where the highest amount of core was added, as majority of the S precursor contributes to the nucleation, the width of the NR was mainly determined by the S amount as shown in the feature weights.

Moreover, larger width NRs were obtained with reduced effect from core size due to increased isotropic growth rate. By varying only the core amount from 60 nmol for sample 03, to 10 nmol for sample 28, the aspect ratios were tuned from 6.1 to 30.5 with almost 5 times increase. In contrast, by varying the core absorption edge over the entire visible range from 489 nm for sample 08, to 570 nm for sample 19, the resulting aspect ratios were only changed from 8.2 to 11.5 with less than 2 times increase, further emphasizing the overall importance of core amount in this reaction. On the other hand, the width of NR was more dependent on S amount available. In order to achieve an anisotropic NR growth, the ligand in the reaction bonded stronger on the side of NR than the head because of the affinity toward the exposed facet. S precursor was thermodynamically more favorable to react on the length growth direction. Therefore, effective S precursor available for width growth in the reaction mixture limited its width. Thus, the feature importance obtained from the model was consistent with experimental observations and reaction mechanism, which further validated the accuracy of the data generation and NR dimension prediction.

**Conclusion**

In conclusion, a deep neural network based regression model has been established to predict the length, width, and aspect ratio of CdSe/CdS seeded core shell nanorods.

Starting with a small dataset of 28 samples, we have successfully demonstrated the excellent performance of the model in accurate prediction of experimental parameters with valid data argumentation through SMOTE-REG. Additional experimental data was tested with well predicted results, further validating the effectiveness of the sample generation and developed model. LIME was further applied to interpret the importance of each condition on the size of the NRs, which matched well with reaction mechanism and experimental observations and further proved the validity of the model. The knowledge obtained and model developed for colloidal semiconductor nanorods with small sample set could be transferred in prediction of other wet chemical synthetic nanocrystal systems and open the possibilities for deep neural network in solving problems with small dataset.

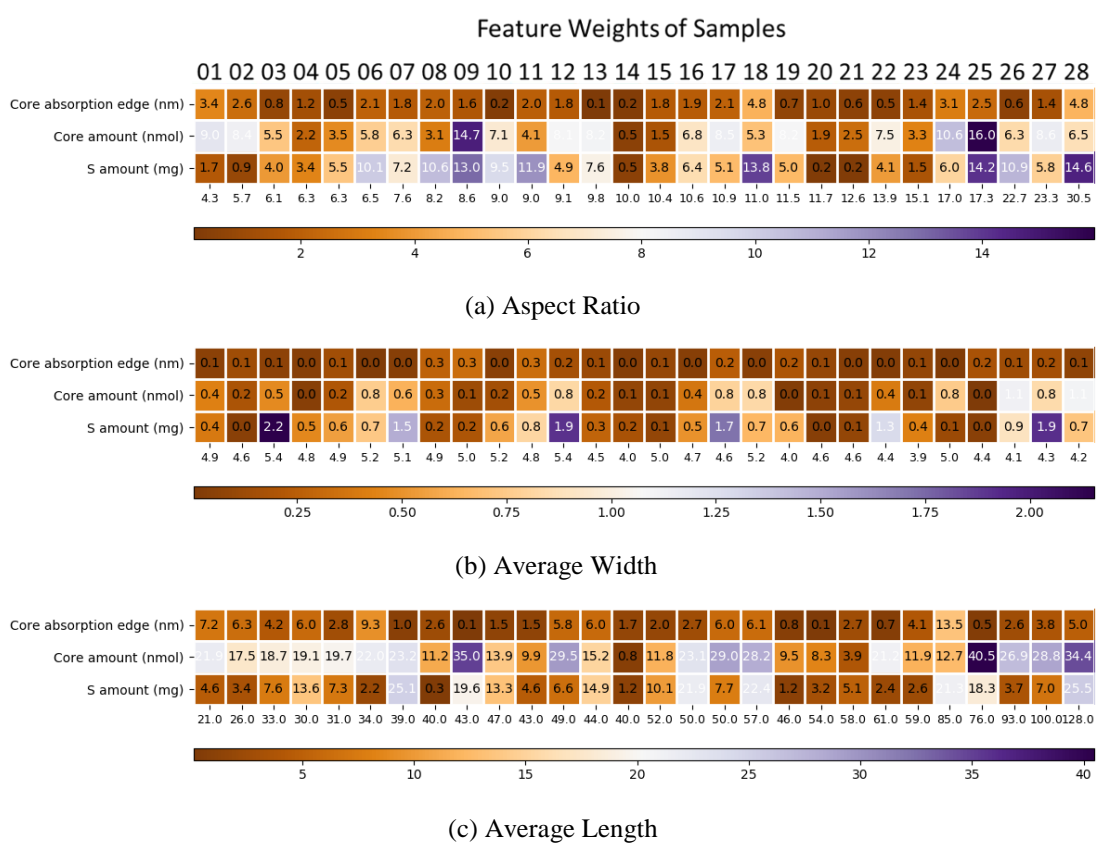

Figure 4 Heatmap of feature weights of samples.

## Methods

In this section, we introduce the three methods used in this work. We first propose the strategy of SMOTE-REG for generating synthetic dataset, which is used to solve the problem that the sample data is too small to train a model accurately. Then, the framework of deep neural network is introduced for the regression task. Finally, we describe LIME method, which can be used to explain the model and the results.

**Synthetic Minority Oversampling Technique (SMOTE)**

SMOTE is an improved scheme based on random oversampling algorithm. Random oversampling adopts the strategy of simple sample replication to increase samples of the minority class, which may easily cause model overfitting. Thus, SMOTE was proposed to analyze the samples of minority class and create artificial examples accordingly. The algorithm flow is as follows and also shown in Figure 5:

(1) For each sample $x$ in minority class, the distance from it to each sample in the same class is calculated with Euclidean distance to obtain its $k$ nearest neighbors.

(2) The sampling ratio $N$ is set according to the unbalanced proportion. For each sample $x$ of minority class, several samples are randomly selected from its $k$ neighbors.

(3) For each randomly selected neighbor $x_n$, a new sample is constructed with the original sample according to Equation (1).

$$x_{new} = x + rand(0,1) \times (\tilde{x} - x) \tag{1}$$

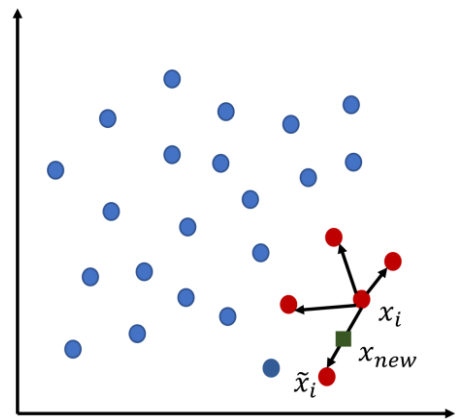

Figure 5 Algorithm flow chart of SMOTE.

The SMOTE method is described in detail below through Algorithm 1.

---

**Algorithm 1** $SMOTE(D, T, N, k)$

Input:
    $T$                                ▷ Number of minority class samples
    $N\%$                              ▷ Amount of SMOTE
    $k$                                   ▷ Number of nearest of neighbors
    $D_{minority}$                   ▷ The original minority class dataset
Output:
    $SyntheticD$                 ▷ Synthetic dataset

1. **begin**
2. **if** $N < 100$ **then**
3.     Randomize the $T$ minority class samples
4.     $D_{minority} \leftarrow$ the first $(N/100) * T$ samples
5.     $T = (N/100) * T$
6.     $N = 100$
7. **end if**
8. $N \leftarrow (int)(N/100)$

```
9.    for i → 1 to T
10.       x̃_i ← k nearest neighbors for x_i ∈ D_{minority}
11.       for j → 1 to N
12.          nn ← a random number between 1 and k
13.          x_i^j = x_i + rand(0,1) × (x̃_{i,nn} − x_i)
14.       end for
15.    end for
16.    SyntheticD ← ⋃_{i∈T, j∈N} x_i^j
17.    return SyntheticD
18. end
```

Furthermore, the proposed SMOTE-REG is described below and in detail through Algorithm 2.

(1) The regression problem is transformed into a classification problem by manually classifying the samples. $D = \{\langle x, y \rangle\}_{i=1}^{N}$ denotes the training sample of the problem. We first manually classify the dataset using the number of minority class samples $T$ to obtain unbalanced subsets. $T$ samples are selected from the sample as a minority class each time, so there are $C_T^N$ combinations. Minority classes are then labeled as Class 1 and the others as Class 2 for each combination, as shown in figure 1  In this way, multiple different minority and majority combinations are generated, that is, an unbalanced subset.

(2) The standard SMOTE algorithm described in Algorithm 1 is applied to each unbalanced data subset. Here, we use SMOTE-REG to get the generated value of $y$ as shown in Equation (2).

$$y_{new} = y + \lambda_x \times (\tilde{y} − y) \qquad (2)$$

where $\lambda_x$ is the random value chosen for the corresponding sample in Equation (1). In this way, we have obtained a new subset of data with a more balanced distribution.

(3) The $C_T^N$ newly generated subsets of data are merged together. The overlapping data in the combined set is eliminated and class labels removed to get the newly generated class-balanced data. For the regression problem, a generated dataset with more samples subject to the sample distribution is then obtained.

The algorithm is described in detail by Algorithm 2 below.

**Algorithm 2** SMOTE for regression algorithm (SMOTE-REG).

```
Input:
     D                              ▷ A regression dataset
     T                              ▷ Number of minority class samples
Output:
     newD                           ▷ Synthetic dataset
1.  begin
```

2. Select T samples as a minority class to generate subset $D_{subset}$ with $C_T^N$ combinations.
3. **for** $D_s$ in $D_{subset}$ **do**
4.     $D_{minority} \leftarrow \{\langle x_i, y_i \rangle \in D_s\}$
5.     $D_{majority} \leftarrow \{\langle x_i, y_i \rangle \notin D_s\}$
6.     $SyntheticD[n] \leftarrow SMOTE(D_{minority}, T, N, k)$
7. **end for**
8.   Combine: $D_{combine} \leftarrow SyntheticD \cup D$
9.   Delete: $newD \leftarrow drop\_duplicates(D_{combine})$
10. **end**

**Deep learning for regression**

In a regression problem, we aim to predict the output of a continuous value, in contrast with a classification problem, where we aim to select a class from a list of classes. The architecture of the neural network[42] used in this work is shown in Figure 6.

The well-trained network had four layers, including an input layer, two fully connected hidden layers, and an output layer that returns a single, continuous value. The activate function used for hidden units is ReLU. Suppose $z^t$ is the data vector in a particular layer $t$, the function of following layer is defined as:

$$z^{t+1} = ReLU(W^t z^t + b^t) = max\{0, W^t z^t + b^t\} \qquad (3)$$

where $W^t$ is the synaptic weight matrices and $b^t$ is the bias vector.

**Local Interpretable Model-Agnostic Explanations (LIME)**

LIME is by far the most widely used explanatory model, which provides a local linear approximation of the behavior of the model.[42,43] While the model may be very complex globally, it is easier to approximate it around the vicinity of a particular instance. Treating the model as a black box, LIME uses a linear model to fit the local behavior of the original model, so that the linear model gives the relative importance of the different features in the sample. Specifically, LIME learns a sparse linear model around the selected point as an explanation by perturbing the instance to be explained.

In order to be independent of the model, this method does not go deep inside the model. To figure out which part of the input contributes to the prediction results, LIME makes small perturbations around the input values to observe the prediction behavior of the model. Let $x$ denote the input interpretation sample and $x_n$ be the sample instances around $x$ in a small range. $x_n$ is entered into model $f(x)$ to be interpreted to get the predicted value of model $f(x_n)$. In this way, a new perturbation dataset $D_n = \{\langle x_n, f(x_n) \rangle\}$ is formed. Then weights are assigned based on the distances of the perturbed data points from the original data, based on which they learn an interpretable model and predictions. The procedure is illustrated in Figure 7, where the light green and yellow areas represent two classes predicted by the black-box model $f$. The dark green and red points are samples of two classes predicted by $f$ and the red cross is the instance to be explained. LIME first generates plenty of samples by linear interpolation in a small neighborhood of this instance. The linear model is then trained using the generated samples to obtain the blue dashed line in the figure. It is worth to note that this line can only represent locally faithful explanation of the model but not globally.

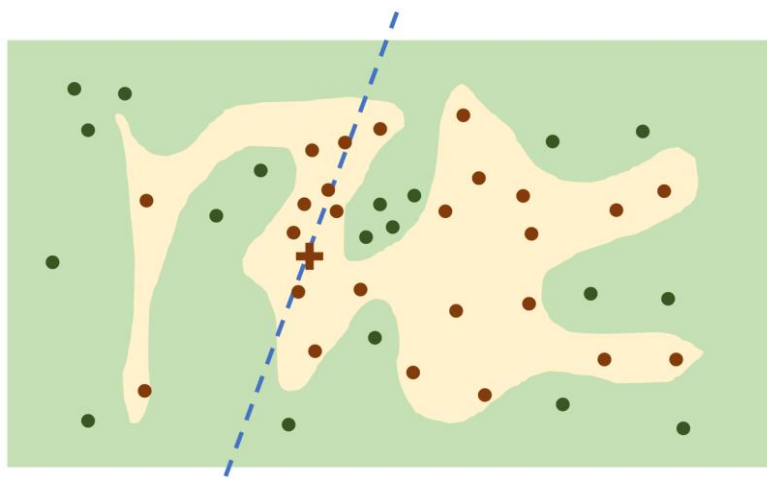

Figure 7 Illustration of LIME method.

**Data and Code Availability**
The datasets used during the current study are available from the corresponding author on reasonable request.